\documentclass[10pt,twocolumn,letterpaper]{article}

\usepackage[pagenumbers]{cvpr}
\usepackage{float}
\usepackage{stfloats}
\usepackage{marvosym}

\definecolor{cvprblue}{rgb}{0.21,0.49,0.74}
\usepackage[pagebackref,breaklinks,colorlinks,allcolors=cvprblue]{hyperref}

\def\ours{{NUMINA}}
\usepackage{colortbl}

\definecolor{linecolor}{rgb}{0.82, 0.94, 0.75}

\definecolor{lowred}{RGB}{238,18,137}

\newcommand{\dplus}[1]{\fontsize{6pt}{0.1em}\selectfont (\textbf{\textcolor{lowred}{#1}})}
\definecolor{evaunit01green}{RGB}{54,125,189}
\newcommand{\evagreen}[1]{\textcolor{evaunit01green}{#1}}
\newcommand{\dtplus}[1]{\fontsize{6pt}{0.1em}\selectfont (\textbf{\evagreen{#1}})}
\definecolor{equalblack}{RGB}{0,0,0}

\newcommand{\equal}[1]{\fontsize{6pt}{0.1em}\selectfont (\textbf{\textcolor{equalblack}{#1}})}

\def\confName{CVPR}
\def\confYear{2026}

\title{When Numbers Speak: Aligning Textual Numerals and Visual Instances \\in Text-to-Video Diffusion Models}

\author{
    Zhengyang Sun$^{1*}$,
    Yu Chen$^{1*}$, 
    Xin Zhou$^{1,3}$, 
    Xiaofan Li$^{2}$, 
    Xiwu Chen$^{3^{\dag}}$, 
    Dingkang Liang$^{1^{\dag}}$, 
    Xiang Bai$^{1\text{\Letter}}$\\
    $^{1}$ Huazhong University of Science and Technology,
    $^{2}$ Zhejiang University,
    $^{3}$ Afari Intelligent Drive
    \\{\tt\small\{zysun,yuchen66,dkliang\}@hust.edu.cn}
    \\{\small Project webpage: \url{https://h-embodvis.github.io/NUMINA/}}
    }

\begin{document}

\twocolumn[
\maketitle

{%
\begin{figure}[H]
\vspace{-40pt}
\hsize=\textwidth
\centering
\includegraphics[width=1.98\linewidth]{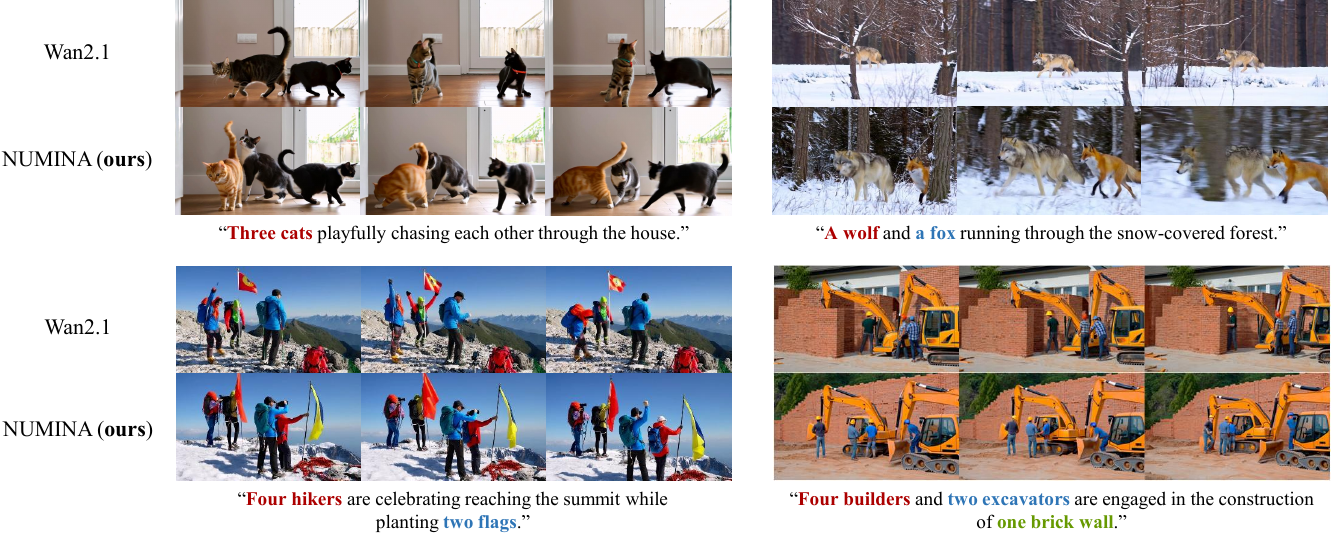}
\vspace{-5pt}
\caption{We present \textbf{\ours}, a training-free framework that alleviates the misalignment between precise numerals and visual instances in text-to-video diffusion models. We significantly improve counting accuracy while maintaining natural layouts and temporal coherence.}
\vspace{-0.5mm}
\label{fig:intro_vis}
\end{figure}
}]

{\let\thefootnote\relax\footnotetext{\hspace{-1.5em}* Equal contribution. $\dag$ Project lead. $\text{\Letter}$ Corresponding author.}}

\begin{abstract}
Text-to-video diffusion models have enabled open-ended video synthesis, but often struggle with generating the correct number of objects specified in a prompt. We introduce \ours~, a training-free identify-then-guide framework for improved numerical alignment. \ours~identifies prompt-layout inconsistencies by selecting discriminative self- and cross-attention heads to derive a countable latent layout. It then refines this layout conservatively and modulates cross-attention to guide regeneration. On the introduced CountBench, \ours~improves counting accuracy by up to 7.4\% on Wan2.1-1.3B, and by 4.9\% and 5.5\% on 5B and 14B models, respectively. Furthermore, CLIP alignment is improved while maintaining temporal consistency. These results demonstrate that structural guidance complements seed search and prompt enhancement, offering a practical path toward count-accurate text-to-video diffusion. The code is available at \url{https://github.com/H-EmbodVis/NUMINA}.

\end{abstract}

\section{Introduction}
\label{sec:intro}

Recent advances in text-to-video (T2V) models~\cite{blattmann2023stable,chen2024videocrafter2,bar2024lumiere,li2025fvar} have greatly enhanced the ability to generate coherent and high-quality videos from textual descriptions. This progress is largely facilitated by the Diffusion Transformer (DiT) architecture~\cite{peebles2023scalable}, enabling scalable training and stronger semantic alignment. By making high-quality video creation more accessible, these models enable emerging applications across entertainment, education, and other domains.

Despite the significant progress, most state-of-the-art T2V models~\cite{li2024hunyuan,wan2025} primarily emphasize visual fidelity~\cite{fang2025panoramic,chen2024gentron}, motion smoothness~\cite{wu2024motionbooth,yang2024motion}, and overall semantic alignment~\cite{gao2025devil,kim2025free2guide}. However, they often struggle with precise numerical alignment between prompts and objects, as shown in Fig.~\ref{fig:intro_vis}. This limitation, where models fail to represent object counts accurately, hinders their reliability in precision-sensitive applications like instructional visualization. This naturally raises a question: \textit{what prevents T2V models from achieving precise numerical alignment?}

\begin{figure}[t]
  \centering
    \includegraphics[width=1.0\linewidth]{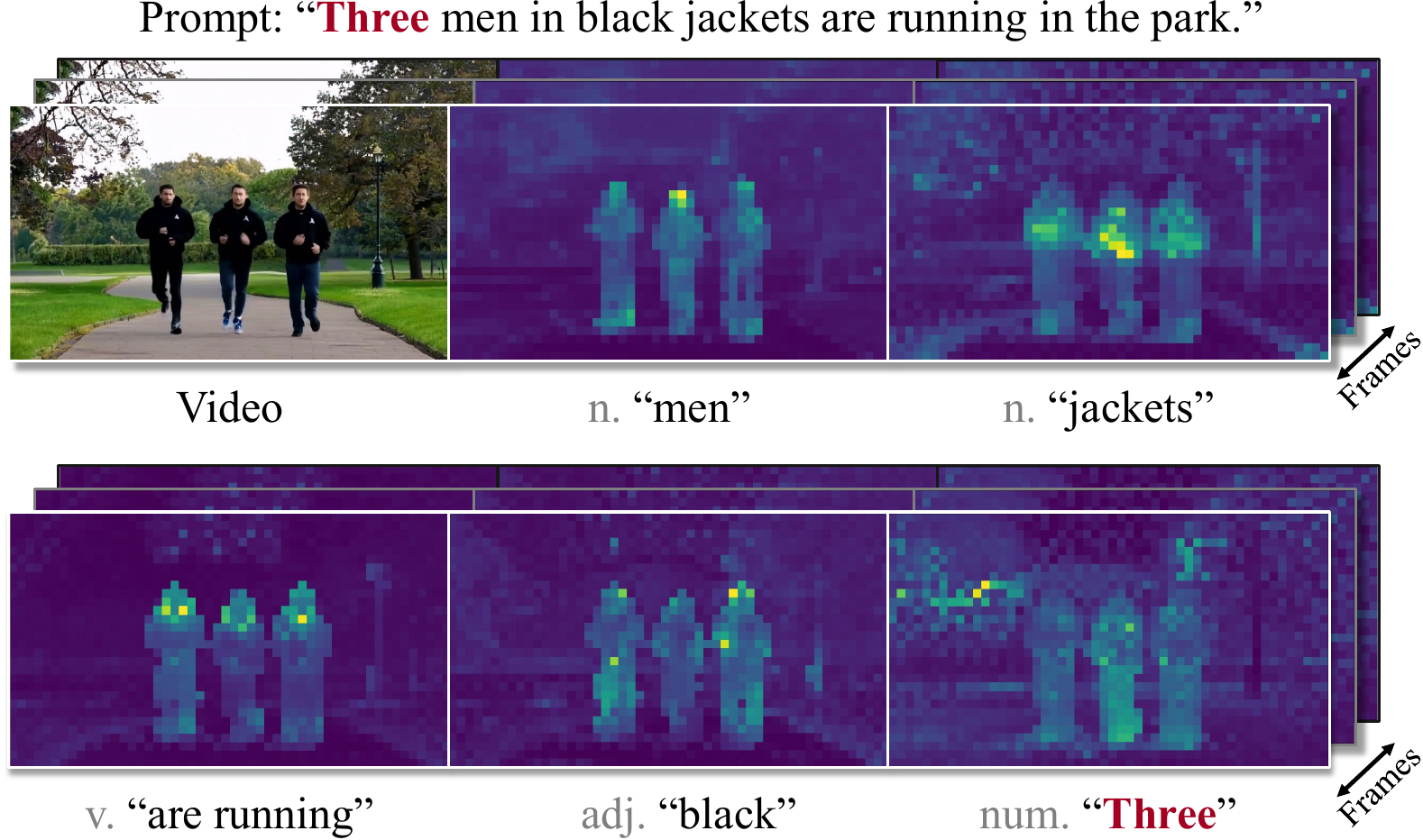}
    \vspace{-19pt}
    \caption{Visualization of the cross-attention maps corresponding to different texts in the prompt. The highlighted areas represent a stronger level of attention between the pixels and the text.}
    \label{fig:intro_crossattn}
    \vspace{-9pt}
\end{figure}

To probe this limitation, we analyze Wan2.1-1.3B~\cite{wan2025}, a representative and community-recognized T2V model, and identify two contributing factors: \textbf{1) Semantic weakness.} Numerical tokens exhibit diffuse cross-attention responses compared to other word types. As shown in Fig.~\ref{fig:intro_crossattn}, the cross-attention maps for nouns, verbs, and adjectives produce strong, localized patterns. This suggests insufficient semantic grounding of numerals in the latent space, weakening the model's ability to encode count constraints during generation. \textbf{2) Instance ambiguity.} The heavily down-sampled spatiotemporal latent space in DiT-based architectures~\cite{hacohen2024ltx,wang2025towards} limits the separability of individual object representations, making stable count control difficult. While retraining could potentially address these issues, it is computationally prohibitive and requires large-scale datasets with precise numerical annotations, which are non-trivial to construct. Moreover, enhancing the model's attention to numerical tokens demands careful rebalancing of the attention mechanism to maintain performance on other critical attributes such as visual quality and motion coherence. These constraints motivate us to pursue alternative solutions to enhance numerical alignment during generation.

Therefore, we propose \textbf{\ours}, a training-free video generation framework that enhances numerical alignment in T2V generation while preserving visual fidelity and temporal coherence, as shown in Fig.~\ref{fig:intro_vis}. We explore the model's latent ability to separate object instances, while allowing natural instance-level addition and removal. \ours~introduces an identify-then-guide paradigm, which yields accurate cardinalities and retains appearance, motion, and semantics. As the intervention is lightweight and training-free, it is broadly applicable across various T2V models.

Specifically, in the first phase, \ours~operates early during denoising to detect misalignment between numeral tokens and the evolving latent layout (i.e., the spatial distribution of object-related activations). It performs a dynamic selection of attention heads using an object discriminability criterion, then applies a cluster-based algorithm to obtain precise segmentation. In the second phase, \ours~utilizes targeted adjustments to refine the latent layout under count constraints, heuristically considering spatial relationships between instances. The subsequent regeneration process is guided by this adjusted layout, improving count accuracy without degrading non-numerical attributes.

To systematically evaluate \ours, we also introduce the CountBench benchmark, comprising 210 prompts covering counts from 1-8 for scenes involving 1-3 object categories. On CountBench, \ours~improves by 7.4\% counting accuracy on Wan2.1-1.3B~\cite{wan2025} and by 5.5\% on a larger 14B model. Moreover, we observe a consistent increase in CLIP score for various baselines, suggesting that enforcing correct instance counts strengthens overall text-video alignment and yields cleaner scene layouts. The successful integration with inference acceleration techniques like EasyCache~\cite{zhou2025easycache} also reduces inference overhead.

Our major contributions can be summarized as follows: \textbf{1)} We reveal that the attentions in T2V models expose critical visual information related to the number of instances. \textbf{2)} We introduce a training-free framework that guides modifications during generation, enhancing the alignment between object counts and prompt instructions. \textbf{3)} We demonstrate that \ours~advances count-accurate text-to-video generation, highlighting its effectiveness and practicality.

\section{Related Work}

\subsection{Diffusion Transformer for Video Generation}
Text-to-video (T2V) generation has rapidly progressed from early 3D U-Net architectures~\cite{ho2022video,ho2022imagen,chen2023videocrafter1,li2024drivingdiffusion} to scalable Diffusion Transformer (DiT) frameworks~\cite{peebles2023scalable,ma2024latte}. Built on DiT~\cite{yin2025towards}, leading video generation models~\cite{hong2022cogvideo,kong2025hunyuanvideo,wan2025,li2025driverse} have achieved remarkable capabilities in synthesizing coherent, high-fidelity videos. They effectively inject textual semantics via attention mechanisms and operate in compressed latent spaces for efficiency~\cite{hacohen2024ltx,blattmann2023align}. Despite these advantages, current T2V models often exhibit weak grounding of textual numerals and insufficient instance separability, resulting in numerical misalignments during generation.

\begin{figure*}[t]
  \centering
   \includegraphics[width=0.99\linewidth]{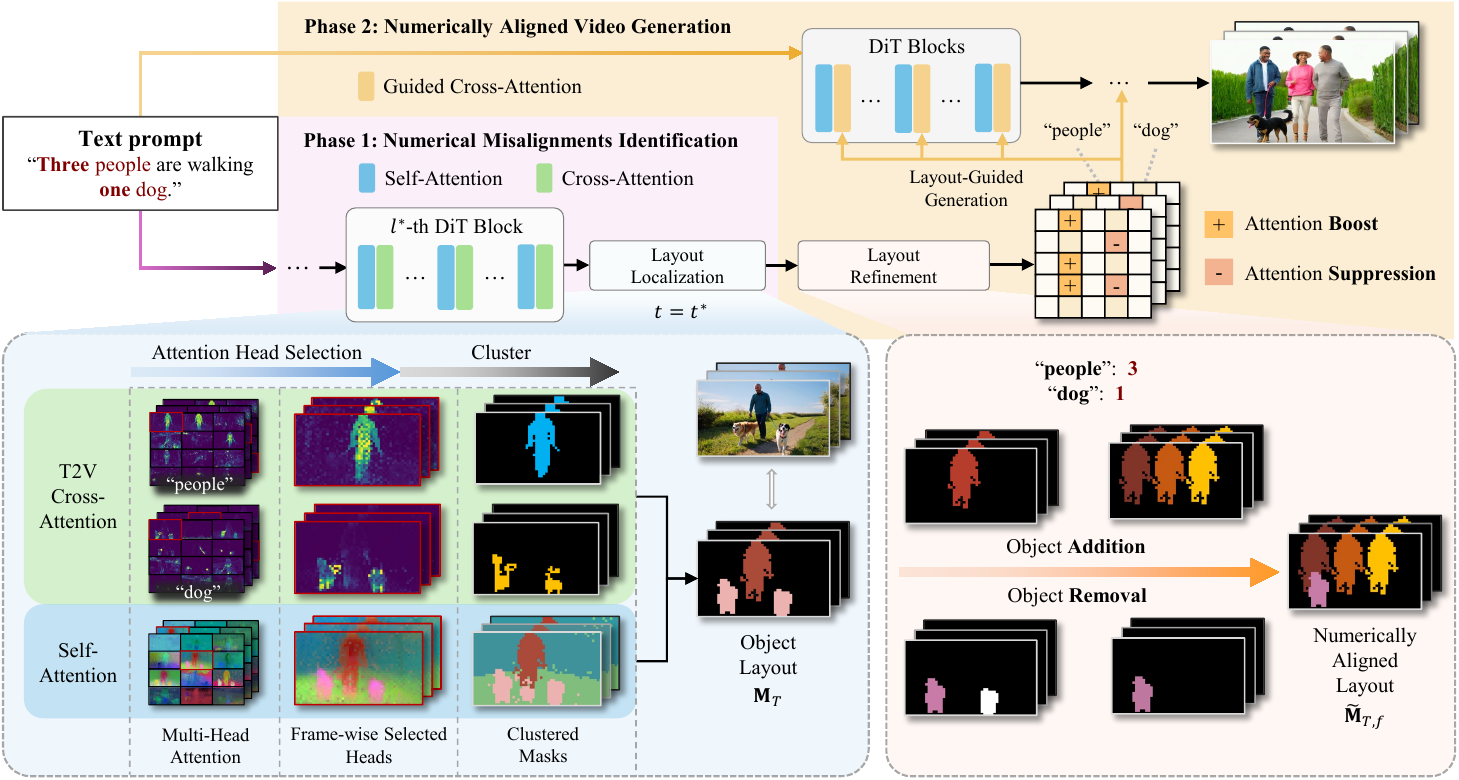}
   \caption{The pipeline of our \ours~follows a two-phase paradigm. Given a text prompt containing numerals, we first perform the numerical misalignment identification to extract explicitly countable layouts from attention maps. Based on the layout, we further conduct a refinement and a layout-guided generation for the numerically aligned video generation.}
   \label{fig:pipeline}
\end{figure*}

\subsection{Video Editing for T2V Models}
Recent advances in T2V models have catalyzed a surge in video editing methods~\cite{jeong2023ground,wu2023tune,bai2025uniedit,lee2025generative,li2025video4edit,yang2025dualdiff+}. Existing approaches predominantly focus on motion control~\cite{gu2024videoswap,molad2023dreamix,wu2024draganything}, style transfer~\cite{yang2022vtoonify,ye2025stylemaster,wang2023videocomposer}, appearance editing~\cite{mou2024revideo,qi2023fatezero,liu2024video}, etc. For example, VideoGrain~\cite{yang2025videograin} supports multi-region and multi-grained editing conditioned on prompts via an attention modulation. Meanwhile, some researchers focus on video inpainting~\cite{thiry2024towards,ding2025homogen,guo2025keyframe}. OmnimatteZero~\cite{samuel2025omnimattezero} and DiffuEraser~\cite{li2025diffueraser} remove objects and their associated visual effects via video inpainting. However, these methods overlook instance-level addition, failing to align textual numerals with visual content. Moreover, most methods operate in a video-to-video setting and rely on object masks obtained from segmentation models~\cite{cheng2023segment,kirillov2023segment}.

\subsection{Counting in Vision and Generation}
While attention mechanisms and vision-language alignment have proven effective for object counting and localization~\cite{liang2022transcrowd,liang2023crowdclip,liang2022focal,liang2025sood++,liang2022end}, enforcing such precise numerical constraints in generative models remains challenging. Recently, CountGen~\cite{binyamin2025countgen} attempts to optimize count-correct text-to-image (T2I) generation by detecting miscounts and employing a learned layout-completion model. However, it is primarily designed for static images, relies on SDXL-specific observations, and requires training additional networks alongside explicit masks for inference-time optimization.

In comparison, our training-free approach provides global guidance for T2V models without requiring input videos, spatial masks, or auxiliary re-layout networks. Importantly, it readily adapts to text-to-video generation while preserving strict temporal consistency.

\section{Preliminary}

Recent text-to-video (T2V) models~\cite{li2024hunyuan,wan2025,cai2025ditctrl} mainly employ the Diffusion Transformer (DiT)~\cite{peebles2023scalable} together with flow-matching~\cite{lipman2023flow,liu2023flow} or diffusion sampling~\cite{nichol2021improved,peebles2023scalable} pipelines that evolve Gaussian noise into a video latent conditioned on a text prompt. In each vanilla DiT block, the prompt is injected mainly through the multi-head cross-attention mechanism. Given spatiotemporal latent features $\mathbf{X}\in\mathbb{R}^{N \times d}$, and text embeddings $\mathbf{c}\in \mathbb{R}^{L \times d}$, the head-wise cross-attention for head $h$ is computed as follows:
\begin{equation}
\mathbf{C}_h= \text{Softmax}\left( \frac{\mathbf{QK}^\top}{\sqrt{d_h}} \right),
\end{equation}
where $\mathbf{Q}$ is projected from the visual latent $\mathbf{X}$, $\mathbf{K}$ comes from text embeddings $\mathbf{c}$, and $d_h=d/n$ is the per-head dimension. The resulting attention map $\mathbf{C}_h \in \mathbb{R}^{N\times L}$ encodes the relevance between each visual and text token.

The cross-attention mechanism is effective for localized attributes due to its per-query matching, but struggles with global constraints. As a result, numeral tokens often exhibit diffuse, low-contrast activations, as in Fig.~\ref{fig:intro_crossattn}. This suggests that standard cross-attention alone may be insufficient to faithfully encode global numerical constraints, implying that simply increasing training data or model size may not be sufficient to address the problem.

In this paper, we alleviate this gap by introducing a training-free framework that explicitly exposes prompt inconsistencies early in denoising and then corrects them. By extracting an instance-aware layout from attentions and enforcing the desired count during regeneration, we provide global guidance for numerical alignment.

\section{Method}
As shown in Fig.~\ref{fig:pipeline}, we utilize a two-phase pipeline for the training-free framework, following an identify-then-guide paradigm. We first perform a pre-generation step using an input prompt and a sampled noise vector to establish the scene layout localization (Sec.~\ref{sec:INM}). Then, we re-generate numerically aligned video through the modified layout guidance (Sec.~\ref{sec:editing}). Overall, our \ours~transforms implicit attention into an explicit layout signal, guiding the generation process to produce more accurate counts.

\subsection{Numerical Misalignment Identification} \label{sec:INM}
The first phase identifies count discrepancies by analyzing the DiT's attention mechanisms. Since attention patterns are distributed across heads, we select the most instance-discriminative self-attention head and the most text-concentrated cross-attention head, and then fuse their maps to obtain an instance-level layout that is explicitly countable, allowing for direct comparison between the estimated cardinality and the prompted numeral.

\begin{figure}[t]
  \centering
   \includegraphics[width=1\linewidth]{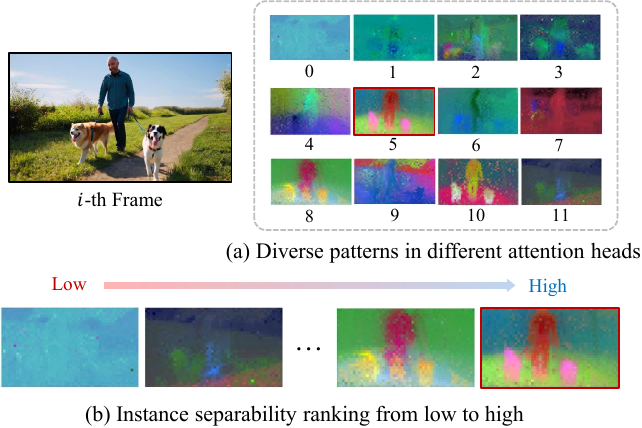}
   \vspace{-15pt}
   \caption{The PCA visualization of self-attention maps for Wan2.1-1.3B. (a) Different attention heads naturally capture diverse spatial patterns. (b) We select the head with the highest instance separability for countable layout construction.}
   \label{fig:sa_pca_vis}
   \vspace{-5pt}
\end{figure}

\textbf{Instance-separable Attention Patterns.} To assess instance awareness, we analyze multi-head attention during early denoising. We observe substantial head-wise diversity in spatial focus, category selectivity, and instance separability. As shown in Fig.~\ref{fig:sa_pca_vis}(a), within the same layer and timestep, many heads are spatially diffuse, some retain coarse class-level structure, and only a small subset clearly delineates boundaries between instances of the same category. This motivates a dynamic head-selection strategy, as naive head averaging or random selection produces blurred maps that fail to separate instances.

\textbf{Attention Head Selection.} Based on these observations, at a reference timestep $t^\star$ during the pre-generation trajectory, we select attention heads from an intermediate layer $\ell^\star$. This balances the emergence of instance contours with the injection of high-level semantics from the prompt, resulting in attention maps with usable structure and limited noise. We process self- and cross-attention separately, due to their distinct roles: self-attention organizes spatial structure, while cross-attention injects prompt semantics. For simplicity, we discuss head selection for a single frame, but the same procedure applies to every latent frame.

To score self-attention heads for instance separability, we first calculate the attention map for each head $h$. For a single frame, we extract the ${HW\times HW}$ normalized attention matrix and apply PCA to project its row vectors onto their top three principal components. The resulting components are reshaped into an ${H\times W\times 3}$ tensor and converted to grayscale, yielding the final map $\mathbf{SA}^{h}\in\mathbb{R}^{H\times W}$ for evaluation. We then design three complementary scores to measure the separability: \textbf{1)} Foreground-background separation $S_{1}^{h}$ is measured by the standard deviation of intensities. \textbf{2)} Structural richness $S_{2}^{h}$ is quantified by partitioning $\mathbf{SA}^{h}$ into non-overlapping blocks, computing the summed feature for each block, and taking the variance across blocks. This metric favors maps with intermediate-scale spatial quality, penalizing both over-smoothing and degeneracy. \textbf{3)} Edge clarity $S_{3}^{h}$ is captured by applying the Sobel operator to $\mathbf{SA}^{h}$ and averaging the gradient magnitude over all pixels, which emphasizes clear object contours that support instance separation. The overall discriminability score for head $h$ is a weighted sum formed as:
\begin{equation}
S\!\left(\mathbf{SA}^{h}\right)=S_{1}^{h}+S_{2}^{h}+\gamma\,S_{3}^{h},
\end{equation}
where $\gamma>0$ balances the contribution of edge clarity against the global contrast and intermediate-scale structure. Finally, as shown in Fig.~\ref{fig:sa_pca_vis}(b), we select the self-attention map $\mathbf{A}_{s}=\mathbf{SA}^{\,h_{s}^{*}}$ by $h_{s}^{*}=\arg\max_{h} S\!\left(\mathbf{SA}^{h}\right)$, providing a layout with highest instance separability.

For each target noun token $T$ in the prompt, we obtain its cross-attention map $\mathbf{CA}^{h}_{T}\in\mathbb{R}^{H\times W}$ from head $h$. We empirically find that the peak activation effectively indicates the model's alignment of the token with a specific visual region. Since these maps are softmax normalized, a higher maximum value $C^{h}_{T}=\max_{x,y}\mathbf{CA}^{h}_{T}(x,y)$ signifies a more concentrated response. For token $T$, we select its best cross-attention head $h_{c}^{*}(T)=\arg\max_{h} C^{h}_{T}$ and denote the corresponding map as $\mathbf{A}_{c,T}=\mathbf{CA}^{\,h_{c}^{*}(T)}_{T}$. This computationally efficient criterion stably identifies relevant regions across scenes without extra processing.

After the selection, we assign one self-attention head $h_{s}^{*}$ for an instance-discriminative spatial scaffold and one cross-attention head $h_{c}^{*}(T)$ per noun token for focused semantic alignment to each frame. These maps jointly yield a countable foreground layout used to compare estimated object cardinality with the prompt numerals.

\textbf{Countable Layout Construction.} To derive a countable layout for a target noun $T$, we fuse the selected instance-discriminative self-attention map $\mathbf{A}_{s}$ and the corresponding text-aligned cross-attention map $\mathbf{A}_{c,T}$ for each frame.

First, spatial proposals $\{\mathbf{r}_i\}$ are generated by partitioning the self-attention map $\mathbf{A}_{s}$ into contiguous regions using clustering~\cite{comaniciu2002mean}. Meanwhile, $\mathbf{A}_{c,T}$ is processed by suppressing values below a 0.1 peak-ratio threshold to isolate peak responses, and density-based clustering~\cite{ester1996density} is then applied to group the resulting map, forming the focus mask $\mathbf{F}$.

We then filter these proposals to construct the final layout. For each region $\mathbf{r}_i$, we compute its semantic overlap score $S_{\text{o}}$ with the focus mask $\mathbf{F}$ as:
\begin{equation} \label{eq:overlap_score}
S_{\text{o}}(\mathbf{r}_i, \mathbf{F}) = \frac{|\mathbf{r}_i \cap \mathbf{F}|}{|\mathbf{r}_i|},
\end{equation}
where $|\cdot|$ denotes the area (number of pixels). A region is retained as a valid instance if $S_{\text{o}} \ge \tau$. The final layout $\mathbf{M}_T$ is then constructed as a 2D semantic map, initialized with a background label $l_{\text{bg}}$. Pixel (with coordinate $p$) belonging to any valid region is assigned the corresponding class $l_T$:
\begin{equation} \label{eq:layout_construction}
\mathbf{M}_T(p) = \begin{cases} l_T, & \text{if } p \in \bigcup_{i: S_{\text{o}}(\mathbf{r}_i, \mathbf{F}) \ge \tau} \mathbf{r}_i \\ l_{\text{bg}}, & \text{otherwise} \end{cases}.
\end{equation}

By construction, $\mathbf{M}_T$ is a semantic map containing disjoint foreground regions, where each region ideally corresponds to a single object instance of category $T$. The number of foreground regions, $|\{i: S_{\text{o}}(\mathbf{r}_i, \mathbf{F}) \ge \tau\}|$, provides an explicit object count. Thus, as in Fig.~\ref{fig:pipeline}, the first stage enables direct identification of numerical misalignments.

\subsection{Numerically Aligned Video Generation} \label{sec:editing}

After identifying numerical misalignment using the layout $\mathbf{M}_T$, this phase alleviates count errors during generation. Since the initial layout $\mathbf{M}_T$ reflects an intrinsic coupling between the sampled noise and the prompt semantics, aggressive manipulation of the latent space can degrade realism. We adopt a conservative two-step approach: Layout Refinement to add or remove object instances at the layout level, and Layout-Guided Generation to steer the re-synthesis process to adhere to this corrected layout.

\textbf{Layout Refinement.}
This process refines the per-frame layout map $\mathbf{M}_{T,f}$ (layout mask of the $f$-th frame for noun $T$) to match the target count $k_T$ parsed from the prompt. Let $m_{T,f}$ be the current number of instance regions in $\mathbf{M}_{T,f}$. The layout is corrected at the instance level until $m_{T,f} = k_T$, guided by a principle of minimal structural change.

For object removal, we erase the smallest region of category $T$ in $\mathbf{M}_{T,f}$ as it incurs minimal perturbation to the existing visual composition.
All pixels within this region are reassigned to the background label. This simple strategy reduces visual impact because small instances carry less spatial support, and it preserves the dominant layout.

For object addition, we insert a new instance using a layout template. If at least one region of category $T$ already exists, the smallest existing region is copied as the template $\mathbf{C}_i$ to preserve the category's intrinsic scale and shape. If no such region exists (i.e., $m_{T,f} = 0$), a circle with radius $r$ is used as $\mathbf{C}_i$. This template defines only the instance's geometry, while the appearance is not constrained.

The template $\mathbf{C}_i$ is then placed at an optimal location in each frame $f$ by minimizing a heuristic cost over a uniform grid of candidate centers. Let $c=(c_x,c_y)$ be the candidate center of $\mathbf{C}_i$, $(c^0_x,c^0_y)$ be the geometric center of $\mathbf{M}_{T,f}$ (which defaults to the spatial center of the frame if $\mathbf{M}_{T,f}$ is empty), and $(c'_x,c'_y)$ be the instance's center in the previous frame. The heuristic cost promotes conservative insertion and is composed of three terms defined as:
\begin{equation}
\begin{aligned}
\mathcal{C}_o &= \big\lvert\,\mathbf{C}_i \cap \mathbf{M}_{T,f}\,\big\rvert, \\
\mathcal{C}_c &= (c_x-c^{0}_x)^2 + (c_y-c^{0}_y)^2, \\
\mathcal{C}_t &= \mathbb{1}[f>1]\big[(c_x-c'_x)^2 + (c_y-c'_y)^2\big],
\end{aligned}
\end{equation}
where $\mathbb{1}[f>1]$ equals 1 when $f>1$ and 0 otherwise. The overlap term $\mathcal{C}_o$ penalizes collisions with the existing layout. The center term $\mathcal{C}_c$ encourages plausible placements close to the existing spatial distribution. The temporal term $\mathcal{C}_t$ ensures the inserted instance remains stable across frames. The total cost $\mathcal{C}$ is a weighted sum as:
\begin{equation}
\mathcal{C}(c)=\mathcal{C}_o+\mathcal{C}_c+\lambda\,\mathcal{C}_t,
\end{equation}
where $\lambda>0$ is a balancing coefficient. The optimal center $c^* = \arg\min_c \mathcal{C}(c)$ is selected, and $\mathbf{M}_{T,f}$ is updated by assigning the class label to the pixels in $\mathbf{C}_{c^*}$. This cycle is repeated until the count $m_{T,f}$ matches $k_T$.

The resulting refined layout $\tilde{\mathbf{M}}_{T,f}$ preserves the original spatial organization while correcting count errors, serving as the control guidance for the subsequent re-generation.

\textbf{Layout-Guided Generation.} Finally, the refined layout $\tilde{\mathbf{M}}_{T,f}$ guides the regeneration process through a training-free modulation of the cross-attention: $\mathrm{softmax}(\mathbf{S}_{\text{pre}} + \mathbf{B})\mathbf{V}$, where $\mathbf{S}_{\text{pre}} = \frac{\mathbf{QK}^\top}{\sqrt{d_h}}$ represents the pre-softmax attention scores and $\mathbf{B}$ is an initially zero bias term. To enforce the corrected layout, we strategically modify either $\mathbf{S}_{\text{pre}}$ or $\mathbf{B}$ for each attention head. These modifications are scaled by a monotonically decreasing intensity function $\delta(t)$ at the $t$-th denoising step, applying stronger guidance early in the denoising process when the object layout is established, and weaker guidance later to preserve fine-grained details.

For object removal, we perform an attention suppression by modifying the bias $\mathbf{B}$ for regions $\Delta\mathbf{M}_{\text{rem}}$ corresponding to category token $T$ to a large negative constant. This forces the post-softmax attention weights in these regions to near zero, effectively suppressing unwanted instance generation.

For object addition, we boost attention in the new area $\Delta\mathbf{M}_{\text{add}}$, and the boost strategy depends on the template's origin. If the new instance is obtained from the manual circle template, we modify the bias term $\mathbf{B}$ by setting it to $k \cdot \delta(t)$ for all $p \in \Delta\mathbf{M}_{\text{add}}$, where $k$ is a scalar coefficient. This provides a strong, externally-imposed attention signal. Conversely, if the instance is templated from an existing reference region $\mathbf{M}_{\text{ref}}$, we directly overwrite the pre-softmax scores in $\mathbf{S}_{\text{pre}}$ to promote consistency. Specifically, for each frame $f$, we compute the mean pre-softmax score $\bar{a}_f$ from $\mathbf{M}_{\text{ref}}$ and then overwrite the scores in $\mathbf{S}_{\text{pre}}$ for all $p \in \Delta\mathbf{M}_{\text{add}}$ at frame $f$ with $\bar{a}_f \cdot \delta(t)$. This transfers the pretrained attention properties of the existing object onto the new location, with $\delta(t)$ modulating the intensity.

This process is applied independently to each category $T$, and the localized guidance ensures stable control superposition while preserving overall visual fidelity.

\section{Experiments}

\subsection{Experiment Setup}
\textbf{Benchmark.} Existing text-to-video (T2V) benchmarks~\cite{liu2024evalcrafter,wu2024towards,huang2024vbench} often overlook precise numerical generation, focusing instead on visual quality~\cite{jayasumana2024rethinking}, temporal coherence~\cite{unterthiner2019fvd}, or general text alignment~\cite{hessel2021clipscore}. While T2VCompBench~\cite{sun2025t2v} includes a numeracy subset, its formulaic structure (``[X] and [Y]") limits its ability to represent diverse user prompts.

To evaluate numerical alignment in T2V generation, we introduce CountBench, comprising 210 prompts that evaluate numerical fidelity in complex scenarios. These prompts encompass a range of conditions, including instance counts from 1 to 8 and compositions involving 1 to 3 object categories, systematically evaluating a T2V model's ability to manage multiple numerical constraints. We initially generated prompt candidates using GPT-5~\cite{openai2025gpt5} to ensure simple and dynamic descriptions, followed by a manual review to eliminate repetitive or illogical prompts.

\textbf{Evaluation metrics.}
We employ three complementary metrics to quantitatively assess numerical alignment and generation quality. \textbf{1)} Counting Accuracy (CountAcc) measures adherence to numerical instructions by scoring a target object class as 1 if the detected count matches the prompt, and 0 otherwise. For each frame, scores are averaged across classes, and then averaged across frames to produce the video-level score. \textbf{2)} Temporal Consistency (TC) measures the stability of generated counts. For each adjacent frame pair, a class scores 1 if counts are identical, and 0 otherwise, with the final score averaged over all pairs and classes. \textbf{3)} The CLIP score~\cite{hessel2021clipscore} evaluates semantic alignment between generated videos and text prompts by averaging frame-wise CLIP scores. The CountAcc and TC are computed using GroundingDINO~\cite{liu2024grounding} to obtain per-frame object counts via category-specific text prompts.

\textbf{Implementation Details.}
We implement \ours~using the official Wan T2V series~\cite{wan2025} with 50 denoising steps. For the numerical misalignment identification stage, we extract attention at timestep $t^\star{=}20$ and layer $\ell^\star{=}15$. All baseline methods share identical inference settings to ensure fair comparison. Experiments on the 1.3B model are conducted on NVIDIA 4090 GPUs, while larger models (5B and 14B) are evaluated on A800 GPUs. 

\begin{table}[!t]
\footnotesize
\setlength{\tabcolsep}{2.25mm}
\centering
\caption{Comparison of \ours~with other practical strategies. We report Counting Accuracy (CountAcc), Temporal Consistency (TC), and CLIP Score on Wan~\cite{wan2025} of varying scales.}
\vspace{-5pt}
\label{tab:main}

\begin{tabular}{ lccccc }
    \toprule
    Models &CountAcc (\%) &TC (\%) &CLIP Score \\
    \midrule
    \multicolumn{4}{c}{Wan2.1-1.3B~\cite{wan2025} (81 frames, 832$\times$480)} \\
    \midrule
    Wan2.1-1.3B &42.3 &81.2 & 33.9\\
    + Seed search &45.5\dplus{+3.2} &82.3\dplus{+1.1} &34.6\dplus{+0.7}\\
    + Prompt enhancement &47.2\dplus{+4.9} &82.1\dplus{+0.9} &33.7\dtplus{-0.2}\\
    \rowcolor{linecolor}+ \ours~ (\textbf{ours}) &\textbf{49.7}\dplus{+7.4} &\textbf{83.4}\dplus{+2.2} & \textbf{35.6}\dplus{+1.7}\\
    \midrule
    \multicolumn{4}{c}{Wan2.2-5B~\cite{wan2025} (81 frames, 1280$\times$704)} \\
    \midrule
    Wan2.2-5B &47.8 & 85.0 &34.3 \\
    + Seed search &48.8\dplus{+1.0} &84.7\dtplus{-0.3} &34.1\dtplus{-0.2} \\
    + Prompt enhancement &49.0\dplus{+1.2} &84.3\dtplus{-0.7} &34.2\dtplus{-0.1}\\
    \rowcolor{linecolor}+ \ours~ (\textbf{ours}) &\textbf{52.7}\dplus{+4.9} &\textbf{85.0}\equal{+0.0} &\textbf{34.7}\dplus{+0.4} \\
    \midrule
    \multicolumn{4}{c}{Wan2.1-14B~\cite{wan2025} (81 frames, 1280$\times$720)} \\
    \midrule
    Wan2.1-14B &53.6 &83.3 &34.2 \\
    + Seed search &56.1\dplus{+2.5} &83.5\dplus{+0.2} & 34.0\dtplus{-0.2} \\
    + Prompt enhancement &56.9\dplus{+3.3} &\textbf{84.3}\dplus{+1.0} &34.0\dtplus{-0.2} \\
    \rowcolor{linecolor}+ \ours~ (\textbf{ours}) &\textbf{59.1}\dplus{+5.5} &84.0\dplus{+0.7} &\textbf{34.4}\dplus{+0.2} \\
    \bottomrule
\end{tabular}
\vspace{-15pt}
\end{table}

\subsection{Main Results}
We conduct experiments on leading video generation models, with Wan models~\cite{wan2025} across different scales, i.e., Wan2.1-1.3B, Wan2.2-5B, Wan2.1-14B, with a fixed seed 1. Since the field of count-aligned T2V generation remains unexplored, we compare our \ours~against the original models and the two most practical and viable training-free strategies within existing generation workflows: \textbf{1) Seed search}, a common ``trial-and-error" practice involving generating 5 videos with random seeds (1-5) per prompt and selecting the best result based on counting accuracy; \textbf{2) Prompt enhancement}, which enriches object descriptions with detailed attributes using a Large Language Model~\cite{anthropic2025claude45sonnet}.

As shown in Tab.~\ref{tab:main}, \ours~consistently and significantly improves counting accuracy (CountAcc) across all baselines. For instance, Wan2.1-1.3B achieves only 42.3\% accuracy with a single trial, while Seed search and Prompt enhancement offer marginal improvements to 45.5\% and 47.2\%, respectively. In contrast, \ours~substantially boosts the accuracy to 49.7\% with only a single trial and a simple prompt. This superior performance extends to larger models, where our method outperforms the 5B and 14B baselines by 4.9\% and 5.5\%, respectively. Notably, our method enables the 1.3B model (49.7\%) to surpass the counting accuracy of the much larger Wan2.2-5B (47.8\%), highlighting its efficiency and effectiveness.

Furthermore, our method improves counting accuracy while maintaining competitive overall generation quality. As shown in Tab.~\ref{tab:main}, we observe a consistent increase in the CLIP score, particularly for smaller models (e.g., an improvement from 33.9 to 35.6 for the 1.3B model). This indicates that enforcing correct spatial layouts and appending missing instances strengthens video-text semantic alignment. Moreover, we find that even state-of-the-art models can exhibit instability in Temporal Consistency (TC). Despite actively adding or removing objects, our method maintains this temporal coherence, and even notably improves it to 84.0\% for the 14B model. This highlights that our instance-level guidance is stable and does not introduce flickering or temporal artifacts, resulting in numerically accurate and temporally smooth videos.

\begin{figure*}[t]
  \centering
   \includegraphics[width=0.9\linewidth]{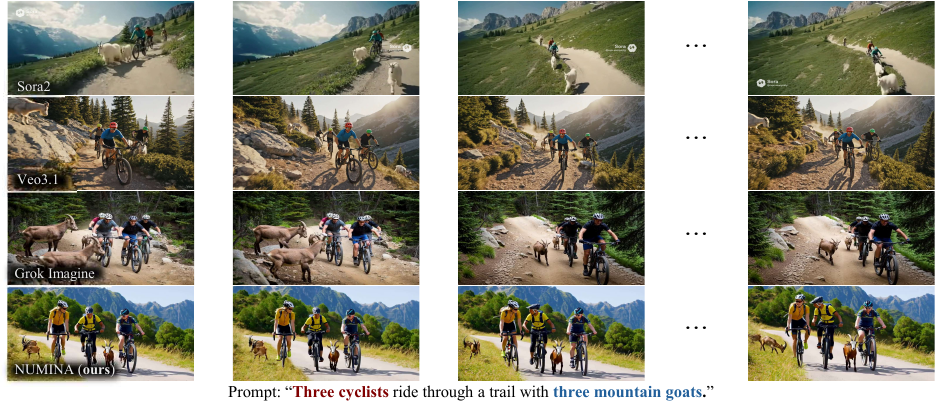}
   \vspace{-7pt}
   \caption{Qualitative comparison of \ours~with the most advanced commercial models.}
   \label{fig:qualitative}
   \vspace{-15pt}
\end{figure*}

\begin{figure}[!t]
  \centering
   \includegraphics[width=0.9\linewidth]{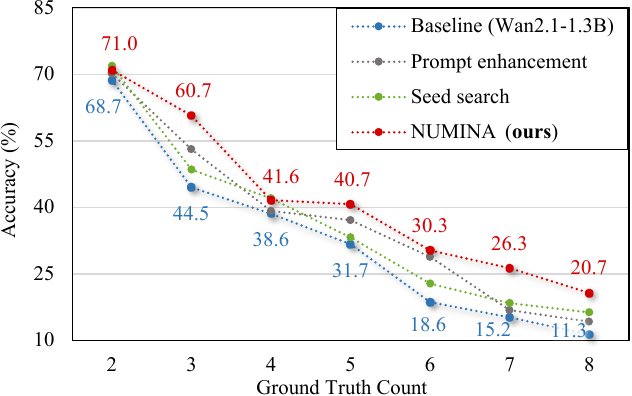}
   \vspace{-10pt}
   \caption{The per-numeral accuracies for Wan2.1-1.3B.}
   \vspace{-10pt}
   \label{fig:acc_count}
\end{figure}

\noindent\textbf{Qualitative Results.} We further present qualitative comparisons with the most advanced commercial T2V generation models in Fig.~\ref{fig:qualitative}. It is worth noting that even these cutting-edge models frequently fail to satisfy the precise numerical constraints specified in the prompt. In contrast, our method reliably produces the exact requested counts while preserving natural layouts and temporal coherence.

\noindent\textbf{Per-numeral Accuracy Breakdown.} Fig.~\ref{fig:acc_count} details a per-numeral breakdown for Wan2.1-1.3B. The 1.3B model already performs well for simple prompts requesting a few objects (e.g., 68.7\% for two objects), as this relies more on category recognition than precise counting. However, its performance rapidly degrades as the ground truth count increases. For prompts requiring three objects, the baseline accuracy plummets to 44.5\%. In contrast, \ours~achieves a 16.2\% improvement, significantly outperforming both Seed search and Prompt enhancement. This advantage is even more pronounced in high-count scenarios. For eight objects, the baseline accuracy is a mere 11.3\%, while \ours~makes a significant improvement by nearly doubling this accuracy to 20.7\%. These results demonstrate that our \ours~provides a scalable solution for complex, high-count scenarios, proving far more effective than augmentation strategies.

\begin{figure}[!t]
  \centering
   \includegraphics[width=0.9\linewidth]{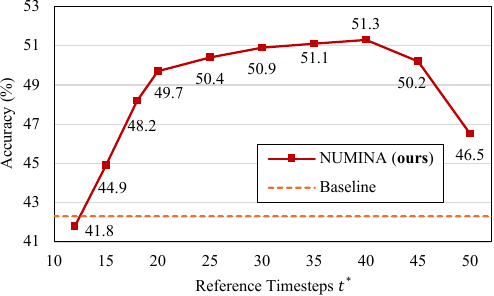}
   \vspace{-10pt}
   \caption{Ablation on the reference timesteps $t^\star$ for head selection.}
   \label{fig:acc_timestep}
   \vspace{-10pt}
\end{figure}

\subsection{Analysis and Ablation Study}
We conduct ablation studies using CountBench prompts, each generating one video with Wan2.1-1.3B unless otherwise specified. The default settings are marked in \colorbox{linecolor}{green}.

\noindent\textbf{Analysis on reference timesteps.} We analyze the impact of the reference timestep $t^\star$ for attention head selection. As shown in Fig.~\ref{fig:acc_timestep}, the CountAcc rises quickly and reaches $49.7\%$ at timestep 20, indicating that early denoising steps are needed to form instance-separable attention. Increasing $t^\star$ to 40 yields only a $3.2\%$ relative gain over $t^\star{=}20$ but doubles the pre-generation cost. For $t^\star{>}40$, we observe an accuracy decline, possibly due to fragmented or over-fused late-stage attention reducing separability. We set $t^\star{=}20$ by default as a favorable accuracy-efficiency trade-off.

\begin{table}[!t]
\footnotesize
\setlength{\tabcolsep}{6.2mm}
\centering
\caption{Ablation on the layout construction method.}
\vspace{-10pt}
\label{tab:dino}
\begin{tabular}{ lcc }
    \toprule
    Method &CountAcc (\%) &TC (\%) \\
    \midrule
    Baseline &42.3 &81.2\\
    \midrule
    GroundingDINO &47.5\dplus{+5.2} &82.8\dplus{+1.6}\\
    \rowcolor{linecolor}Attention (\textbf{ours}) &\textbf{49.7}\dplus{+7.4} &\textbf{83.4}\dplus{+2.2}\\
    \bottomrule
\end{tabular}
\vspace{-5pt}
\end{table}

\begin{table}[!t]
\footnotesize
\setlength{\tabcolsep}{4.7mm}
\centering
\caption{Ablation on the components of the layout refinement cost.}
\vspace{-10pt}
\label{tab:refinement_cost}
\begin{tabular}{ ccccc }
    \toprule
    $\mathcal{C}_o$ & $\mathcal{C}_c$ & $\mathcal{C}_t$ & CountAcc (\%) & TC (\%) \\
    \midrule
    \multicolumn{3}{c}{Baseline} & 42.3 & 81.2 \\
    \midrule
    $\checkmark$ & & & 45.1\dplus{+2.8} & 82.1\dplus{+0.9} \\
    $\checkmark$ & $\checkmark$ & & 46.9\dplus{+4.6} & 82.3\dplus{+1.1} \\
    $\checkmark$ & & $\checkmark$ & 48.9\dplus{+6.6} & 83.1\dplus{+1.9} \\
    \rowcolor{linecolor}$\checkmark$ & $\checkmark$ & $\checkmark$ & \textbf{49.7}\dplus{+7.4} & \textbf{83.4}\dplus{+2.2} \\
    \bottomrule
\vspace{-6pt}
\end{tabular}
\end{table}

\noindent\textbf{Efficacy of countable layout construction.} Tab.~\ref{tab:dino} presents the quality of the countable layout $\mathbf{M}_T$. For fairness, we perform a truncated pre-generation with $t^\star{=}20$. We then derive our layout from selected self-/cross-attention heads and, in parallel, use GroundingDINO~\cite{liu2024grounding} on the same frames to generate a per-category layout. Both layouts are used in the second phase. Our attention-derived layout outperforms the detector-derived layout by 2.2\%, likely because it is native to the DiT's latent and better captures nascent instance structures. More importantly, both layout-guided methods substantially outperform the baseline, validating the effectiveness of our Layout-Guided strategy.

\noindent\textbf{Analysis on the layout refinement cost.} We also assess the components of our layout refinement cost, which are designed to guide object addition. As shown in Tab.~\ref{tab:refinement_cost}, using only the primary overlap cost ($\mathcal{C}_o$) brings a promising 2.8\% accuracy improvement, demonstrating the layout-guided approach's effectiveness. Building on this, adding the center cost ($\mathcal{C}_c$) for plausible spatial placement further improves accuracy to 46.9\%. Meanwhile, the temporal cost ($\mathcal{C}_t$) yields a more substantial gain to 48.9\%, highlighting the importance of temporal stability. Combining all three costs in \ours~achieves the highest accuracy of 49.7\%, confirming that these heuristic costs are complementary and enable stable and accurate layout correction $\tilde{\mathbf{M}}_{T,f}$.

\noindent\textbf{Analysis on the self-attention head selection.} We then validate our strategy of selecting the single best self-attention head using the score $S(\mathbf{SA}^h)$. As shown in Tab.~\ref{tab:head_select}, selecting a single random head (44.1\%) or averaging all heads (43.0\%) provides only a marginal benefit over the baseline. In contrast, our Top-1 selection based on $S(\mathbf{SA}^h)$ significantly boosts accuracy to 49.7\%. This demonstrates that our scoring metric is highly effective at identifying the most useful head. Furthermore, performance consistently degrades as we average fewer discriminative heads, strongly confirming our hypothesis that instance-separable information is a sparse property held by only a few heads.

\noindent\textbf{Analysis of computational overhead.}
Besides, \ours~is compatible with inference acceleration techniques like EasyCache~\cite{zhou2025easycache}, as the pre-processing stage focuses on creating a coarse latent layout, avoiding the need for high-precision computation. As shown in Tab.~\ref{tab:time}, this integration effectively reduces pre-processing overhead with minimal VRAM usage and acceptable wall-clock time. This accelerated pipeline offers a highly efficient and deterministic alternative to the exhaustive seed search typically required for accurate counting.

\begin{table}[t]
\footnotesize
\setlength{\tabcolsep}{7.2mm}
\centering
\caption{Ablation on the self-attention head selection strategy.}
\vspace{-10pt}
\label{tab:head_select}
\begin{tabular}{ lcc }
    \toprule
    Method &CountAcc (\%) &TC (\%) \\
    \midrule
    Baseline &42.3 &81.2\\
    \midrule
    Random &44.1\dplus{+1.8} &82.6\dplus{+1.4} \\
    All average &43.0\dplus{+0.7} &82.4\dplus{+1.2} \\
    Top-3 &48.2\dplus{+5.9} &82.5\dplus{+1.3} \\
    Top-2 &49.4\dplus{+7.1} &83.3\dplus{+2.1} \\
    \rowcolor{linecolor}Top-1 &\textbf{49.7}\dplus{+7.4} &\textbf{83.4}\dplus{+2.2}\\
    \bottomrule
\vspace{-15pt}
\end{tabular}
\end{table}

\begin{table}[t]
\footnotesize
\setlength{\tabcolsep}{0.4mm}
\centering
\caption{Additional time and VRAM cost.}
\vspace{-10pt}
\label{tab:time}
\begin{tabular}{ lccc }
    \toprule
    Method &wall-clock (s) & VRAM (GB) & CountAcc (\%)\\
    \midrule
    Wan2.1-1.3B &292 & 14.3 & 42.3\\
    NUMINA &431 &16.3 & 49.7\\
    NUMINA + EasyCache~\cite{zhou2025easycache} &355 &16.3 & 49.4\\
    \bottomrule
\end{tabular}
\vspace{-2pt}
\end{table}

\section{Conclusion}
This paper presents \ours, a training-free framework for count alignment in text-to-video diffusion. By leveraging instance-separable attention heads in DiTs, \ours~identifies and corrects prompt-layout inconsistencies through explicit layout construction, conservative refinement, and layout-guided generation. \ours~significantly boosts counting accuracy, particularly at higher counts where baselines falter, without sacrificing video quality. These results highlight the value of structural guidance as a complement to existing methods, offering a practical approach to count-accurate text-to-video generation and improving numeral alignment for broader applicability.

\noindent\textbf{Limitations.} While \ours~significantly improves numerical alignment, achieving perfect accuracy across all scenarios remains challenging. Besides, generating very dense instances (e.g., tens or hundreds) remains unexplored. Enabling fully numerically precise video generation for any number is an important direction for future research.

{\small
\bibliographystyle{ieeenat_fullname}
\bibliography{references}
}

\maketitlesupplementary

\setcounter{section}{0}
\renewcommand{\thesection}{S\arabic{section}}

\section{Additional Results}
\label{sec:add_result}

\textbf{Compatibility with CogVideoX~\cite{yang2025cogvideox}.} 
To substantiate the generalizability and robustness of our method beyond a single model architecture, we evaluate our method on CogVideoX-5B, which employs a Multi-Modal Diffusion Transformer (MMDiT). Unlike vanilla DiTs in Wan models~\cite{wan2025}, MMDiT employs a unified global attention mechanism over concatenated visual-textual tokens without a dedicated cross-attention module. To bridge this gap, we adapt our strategy in \cref{sec:INM} of the manuscript by decomposing the unified attention into distinct components. The video-to-video attention is treated as self-attention, while the text-to-video attention sub-matrix is extracted as cross-attention.

As shown in Tab.~\ref{tab:cogvideo}, quantitative results demonstrate a consistent and significant improvement in numerical accuracy when our method is applied to CogVideoX-5B. Specifically, CogVideoX-5B achieves only 40.2\% accuracy under minimal settings, while Seed search and Prompt enhancement provide limited gains of only 2.5\% and 2.3\%, respectively. In contrast, \ours~substantially elevates the performance to 44.4\% using simple prompts and a single generation pass. Furthermore, our method improves overall generation quality, improving the TC and CLIP scores to 80.2\% and 35.4\%, respectively. This successful extension to MMDiT further confirms the general applicability of our training-free approach across different implementations of the architecture.

\textbf{Integration with enhancement strategies.}
As shown in Tab.~\ref{tab:main} of the manuscript, our method alone achieves substantial improvements on CountBench. We further demonstrate that \ours~is fully compatible with prompt enhancement and seed search, which represent the most accessible techniques for boosting counting accuracy. By integrating our method with these enhancement strategies, we achieve the best performance with 54.2\% counting accuracy, reported in Tab.~\ref{tab:combine}. This combined approach significantly surpasses all compared methods, including our standalone \ours~(49.7\%), prompt enhancement (47.2\%), and seed search (45.5\%). In particular, it also enables the 1.3B model to outperform larger baseline models, including Wan2.2-5B at 47.8\% and Wan2.1-14B at 53.6\%. These results establish our approach as a superior alternative to existing workflows, providing a more effective solution for the challenging counting alignment in video generation.

\begin{table}[t]
\footnotesize
\setlength{\tabcolsep}{2.2mm}
\centering
\caption{Evaluation results on CogVideoX~\cite{yang2025cogvideox}.}
\vspace{-10pt}
\label{tab:cogvideo}
\begin{tabular}{ lccc }
    \toprule
    Models &CountAcc (\%) &TC (\%) &CLIP Score \\
    \midrule
    \multicolumn{4}{c}{CogVideoX-5B~\cite{yang2025cogvideox} (81 frames, 1360$\times$768)} \\
    \midrule
    CogVideoX-5B &40.2 &78.1 &34.8 \\
    + Seed search &42.7\dplus{+2.5} &78.3\dplus{+0.2} &34.8\equal{-0.0}\\
    + Prompt enhancement &42.5\dplus{+2.3} &79.0\dplus{+0.9} &34.5\dtplus{-0.3}\\
    \rowcolor{linecolor}+ \ours~(\textbf{ours}) &\textbf{44.4}\dplus{+4.2} &\textbf{80.2}\dplus{+2.1} & \textbf{35.4}\dplus{+0.6}\\
    \bottomrule
\end{tabular}
\end{table}

\begin{table}[!t]
\footnotesize
\setlength{\tabcolsep}{1.6mm}
\centering
\caption{Ablation on combined methods.}
\vspace{-10pt}
\label{tab:combine}
\begin{tabular}{ lccc }
    \toprule
    Models &CountAcc (\%) &TC (\%) &CLIP Score \\
    \midrule
    \multicolumn{4}{c}{Wan2.1-1.3B~\cite{wan2025} (81 frames, 832$\times$480)} \\
    \midrule
    Wan2.1-1.3B &42.3 &81.2 & 33.9\\
    + Seed search &45.5\dplus{+3.2} &82.3\dplus{+1.1} &34.6\dplus{+0.7}\\
    + Prompt enhancement &47.2\dplus{+4.9} &82.1\dplus{+0.9} &33.7\dtplus{-0.2}\\
    + \ours~(\textbf{ours}) &49.7\dplus{+7.4} &83.4\dplus{+2.2} & \textbf{35.6}\dplus{+1.7}\\
    \rowcolor{linecolor}+ Combined method (\textbf{ours}) & \textbf{54.2}\dplus{+11.9} & \textbf{83.6}\dplus{+2.4} &35.5\dplus{+1.6} \\
    \bottomrule
\end{tabular}
\end{table}

\begin{table}[t]
\footnotesize
\setlength{\tabcolsep}{5.7mm}
\centering
\caption{VBench~\cite{huang2024vbench} Subject-Consistency scores. }
\vspace{-10pt}
\label{tab:vbench}
\begin{tabular}{ lcc }
    \toprule
    Models &Baseline &+ \ours~(\textbf{ours}) \\
    \midrule
    Wan2.1-1.3B &83.1 &\textbf{83.6}\dplus{+0.5} \\
    Wan2.1-14B &84.3 &\textbf{84.7}\dplus{+0.4} \\
    Wan2.2-5B &83.4 &\textbf{83.5}\dplus{+0.1} \\
    CogVideoX-5B &84.6 &\textbf{84.6}\equal{+0.0} \\
    \bottomrule
\end{tabular}
\end{table}

\textbf{Evaluation on VBench~\cite{huang2024vbench} metric.}
To assess the temporal stability of the generated object instances, we adopt the Subject-Consistency metric from VBench. For each instance, we extract DINO~\cite{caron2021emerging} features in all frames and compute the cosine similarity with both the first frame and the preceding frame. The two similarities are averaged, and the final video-level score is obtained by averaging over all non-initial frames. We report the mean score across instances. As shown in Tab.~\ref{tab:vbench}, our method achieves competitive performance on this metric, indicating that the edited instances remain temporally stable and visually coherent. This result further validates the reliability of our TC metric, as both measurements capture complementary aspects of temporal coherence. In addition, our counting accuracy follows the Generative Numeracy evaluation protocol in T2V-CompBench~\cite{sun2025t2v}, ensuring that our overall evaluation framework is both consistent and reliable.

\textbf{Analysis on no-reference addition.}
We analyze the effectiveness of adding missing instances when no reference instances are available. This presents a particularly challenging setting where baseline models typically fail to generate the required objects. As shown in Tab.~\ref{tab:circle}, the no-intervention baseline achieves only 48.8\% accuracy without layout refinements in such cases. To address this limitation, we compare two geometric priors for layout refinement: a circular template and a rectangular alternative of equivalent area. Experimental results demonstrate the effectiveness of both strategies, with the rectangular prior reaching 49.5\% accuracy and the circular prior achieving 49.7\%. In practice, we employ the circular prior as described in \cref{sec:editing} of the manuscript. This design minimizes structural assumptions, granting T2V models the flexibility to interpret and form the most contextually plausible objects.

\begin{table}[t]
\footnotesize
\setlength{\tabcolsep}{6.2mm}
\centering
\caption{Ablation on strategy for no-reference addition.}
\vspace{-10pt}
\label{tab:circle}
\begin{tabular}{ lcccc }
    \toprule
    Method &CountAcc (\%) &TC (\%) \\
    \midrule
    Baseline &42.3 &81.2\\
    \midrule
    No intervention &48.8\dplus{+6.5} &83.0\dplus{+1.8}\\
    Rectangle &49.5\dplus{+7.2} &83.3\dplus{+2.1}\\
    \rowcolor{linecolor}Circle &\textbf{49.7}\dplus{+7.4} &\textbf{83.4}\dplus{+2.2}\\
    \bottomrule
\end{tabular}
\vspace{-5pt}
\end{table}

\begin{figure}[t]
  \centering
   \includegraphics[width=1\linewidth]{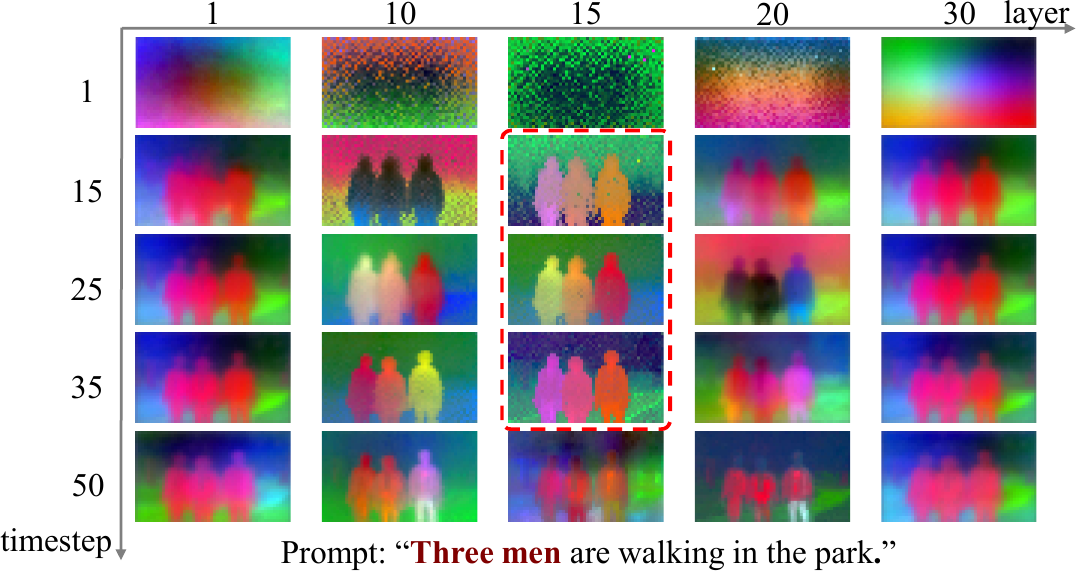}
   \vspace{-15pt}
   \caption{PCA visualization across timesteps and layers.}
   \label{fig:sup_pcavis}
   \vspace{-10pt}
\end{figure}

\textbf{Analysis on layout localization.}
We next analyze the feasibility of layout localization based on Wan2.1-1.3B~\cite{wan2025}. As visualized in Fig.~\ref{fig:sup_pcavis}, our analysis reveals clear instance-separable attention patterns during denoising. These discriminative layouts emerge most distinctly at middle denoising steps, with intermediate layers providing the sharpest spatial separation of object instances. We accordingly set $t^\star=20$ and $\ell^\star=15$ to balance efficiency and accuracy. By performing layout localization at this point and early stopping, we reduce the denoising steps for pre-generation by approximately 60\% without significantly sacrificing accuracy, as quantified in Fig.~\ref{fig:acc_timestep} of the manuscript. This early termination delivers significant computational savings, particularly for larger models. The same relative proportions can be directly applied to other model architectures through straightforward scaling.

\textbf{Analysis on hyperparameters.}
We emphasize that our hyperparameters are generic and are largely set without exhaustive tuning. Selections of layer and timestep vary solely due to intrinsic model differences (e.g., the total number of inference steps) rather than specific heuristic design. We uniformly set $t^\star=20$ and $\ell^\star=15$ in this section for a fair ablation study. As detailed in Tab.~\ref{tab:hyperpara}, our method maintains stable performance across a wide range of hyperparameter values.

\begin{table}[t]
\footnotesize
\centering
\caption{Ablation results for different hyperparameter values.}
\vspace{-10pt}
\label{tab:hyperpara}
\setlength{\tabcolsep}{11pt}
    \begin{tabular}{ ccc }
        \toprule
            $\lambda$ / CountAcc (\%) &$\tau$ / CountAcc (\%) & $k$ / CountAcc (\%)\\
        \midrule
            4 / 49.3  &0.1 / 48.4 &0.5 / 48.2\\
            8 / 49.7  &0.2 / 49.7 &0.8 / 49.7\\
            16 / 49.5 &0.3 / 49.2 &1.0 / 49.2\\
        \bottomrule
    \end{tabular}
\vspace{-5pt}
\end{table}

\textbf{Analysis on the object addition/removal.} We finally analyze the effect of layout-guided generation operations, i.e., object addition and removal. Tab.~\ref{tab:addition_removal} shows that addition alone significantly boosts accuracy by 5.4\%, while removal yields a smaller 1.5\% gain. This suggests that the baseline model primarily struggles with object omission, making addition the more impactful correction. Furthermore, combining both operations achieves the highest accuracy, slightly exceeding the sum of individual gains, proving a synergistic effect between the two complementary guidance methods.

\begin{table}[!t]
\footnotesize
\setlength{\tabcolsep}{4.6mm}
\centering
\caption{Ablation on object addition or removal.}
\vspace{-10pt}
\label{tab:addition_removal}
\begin{tabular}{ cccc }
    \toprule
    Addition & Removal & CountAcc (\%) & TC (\%) \\
    \midrule
    \multicolumn{2}{c}{Baseline} & 42.3 & 81.2 \\
    \midrule
    $\checkmark$ & & 47.7\dplus{+5.4} & 83.0\dplus{+1.8} \\
     & $\checkmark$ & 43.8\dplus{+1.5} & 82.4\dplus{+1.2} \\
    \rowcolor{linecolor}$\checkmark$ & $\checkmark$ & \textbf{49.7}\dplus{+7.4} & \textbf{83.4}\dplus{+2.2} \\
    \bottomrule
\end{tabular}
\vspace{-5pt}
\end{table}

\textbf{Evaluation of visual quality.}
We evaluate visual generation quality using VBench (Aesthetic \& Imaging Quality). As shown in Tab.~\ref{tab:quality}, our method maintains comparable or even superior metric scores, introducing no degradation in video generation quality while significantly enhancing numerical alignment, which is further confirmed by the user study, demonstrating the quality of our approach.

\begin{table}[t]
\footnotesize
\setlength{\tabcolsep}{7.5mm}
\centering
\caption{VBench Aesthetic \& Imaging Quality scores.}
\vspace{-10pt}
\label{tab:quality}
\begin{tabular}{lcc}
    \toprule
    Method & Imaging$\uparrow$ & Aesthetic$\uparrow$ \\
    \midrule
    Wan2.1-1.3B & 71.3\%  & 61.5\% \\
    +NUMINA & 70.9\%  & 63.5\% \\
    \bottomrule
\end{tabular}
\vspace{-5pt}
\end{table}

\textbf{User study.}
We conduct a blind user study involving 10 participants  (balanced gender ratio) using 100 pairs of randomly sampled videos. Participants are asked to evaluate both visual quality and instruction following. The results show a 61\% preference for our method versus 39\% for the baseline. This clear preference confirms that our method delivers not only better objective metric performance but also a superior user experience.

\section{More Visualization}
\label{sec:more_vis}

\begin{figure*}[t]
\vspace{-3mm}
\hsize=\textwidth
\centering
\includegraphics[width=1\linewidth]{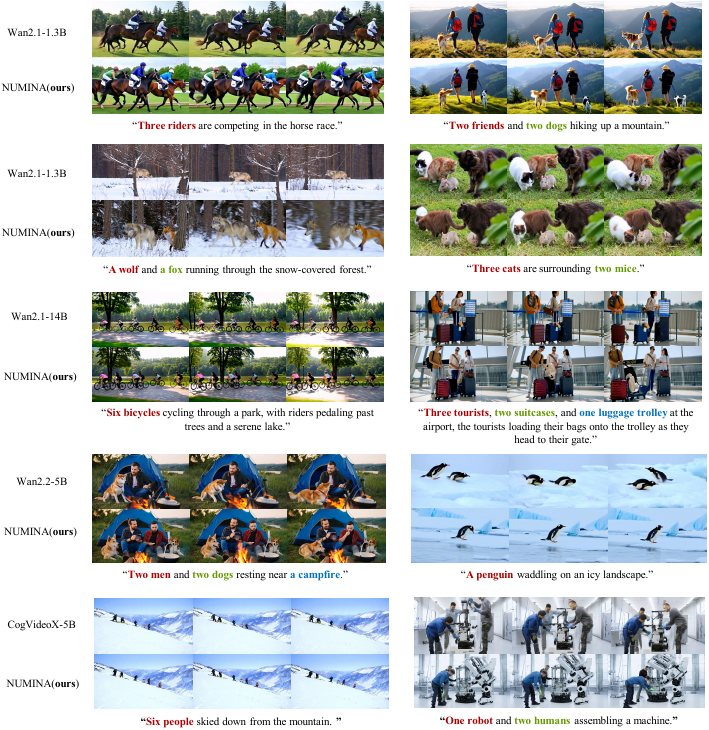}
\vspace{-2pt}
\caption{More representative examples where our method faithfully generates the specified number of objects.}
\vspace{2mm}
\label{fig:more_vis1}
\end{figure*}

\textbf{Additional demos.}
We provide more comprehensive qualitative comparisons in Fig.~\ref{fig:more_vis1}, showcasing our method's effectiveness across different model architectures. The consolidated visualization presents successful numerical alignment cases on Wan2.1~\cite{wan2025} and CogVideoX~\cite{yang2025cogvideox}, demonstrating consistent improvement in generating accurate object counts. These cross-architecture validations collectively confirm our method's strong generalizability and practical utility for enhancing numerical accuracy in text-to-video generation systems. More video demos can be found on our \href{https://h-embodvis.github.io/NUMINA}{project page}.

\begin{figure}[t]
  \centering
   \includegraphics[width=1\linewidth]{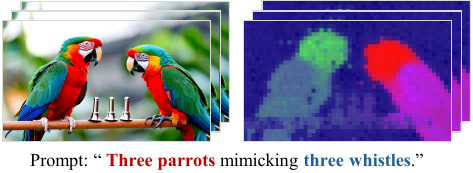}
   \vspace{-15pt}
   \caption{A failure case of \ours. The parrots' heads become decoupled from their bodies in layout construction.}
   \label{fig:failure}
   \vspace{-10pt}
\end{figure}

\textbf{Failure cases.}
A characteristic failure mode of our method occurs when instance-separable attention heads focus excessively on the most salient parts of an object (e.g., an animal's head) rather than its entirety, as demonstrated by the representative failure case in Fig.~\ref{fig:failure}. This leads to an over-segmented layout where parts of a single instance are mistaken for multiple objects, ultimately propagating an irrecoverable error into the final video output. This limitation underscores the challenge of defining instances solely via raw attention and suggests the need for future work to incorporate more holistic perceptual grouping cues.

\end{document}